\documentclass[sigconf,authorversion,nonacm]{acmart}
\usepackage{graphicx}
\usepackage{amsmath}

\usepackage{amssymb}
\usepackage{booktabs}
\usepackage{multirow}
\usepackage{makecell}
\usepackage{indentfirst}
\usepackage{url}
\AtBeginDocument{%
  \providecommand\BibTeX{{%
    \normalfont B\kern-0.5em{\scshape i\kern-0.25em b}\kern-0.8em\TeX}}}

\acmSubmissionID{118}

\begin{document}

\title{RefCrowd: Grounding the Target in Crowd with Referring Expressions}

%
\author{Heqian Qiu, Hongliang Li, Taijin Zhao, }
\author{Lanxiao Wang,Qingbo Wu, Fanman Meng}
\affiliation{%
	\institution{University of Electronic Science and Technology of China}
	\city{Chengdu}
	\country{China}}
\email{hqqiu@std.uestc.edu.cn, hlli@uestc.edu.cn, zhtjww@std.uestc.edu.cn}
\email{lanxiao.wang@std.uestc.edu.cn, qbwu@uestc.edu.cn, fmmeng@uestc.edu.cn}
%
%
%
%
%
%

\renewcommand{\shortauthors}{Trovato and Tobin, et al.}

\begin{abstract}
Crowd understanding has aroused the widespread interest in vision domain due to its important practical significance. Unfortunately, there is no effort to explore crowd understanding in multi-modal domain that bridges natural language and computer vision. Referring expression comprehension (REF) is such a representative multi-modal task. Current REF studies focus more on grounding the target object from multiple distinctive categories in general scenarios. It is difficult to applied to complex real-world crowd understanding. To fill this gap, we propose a new challenging dataset, called RefCrowd, which towards looking for the target person in crowd with referring expressions. It not only requires to sufficiently mine the natural language information, but also requires to carefully focus on subtle differences between the target and a crowd of persons with similar appearance, so as to realize the fine-grained mapping from language to vision. Furthermore, we propose a Fine-grained Multi-modal Attribute Contrastive Network (FMAC) to deal with REF in crowd understanding. It first decomposes the intricate visual and language features into attribute-aware multi-modal features, and then captures discriminative but robustness fine-grained attribute features to effectively distinguish these subtle differences between similar persons. The proposed method outperforms existing state-of-the-art (SoTA) methods on our RefCrowd dataset and existing REF datasets. In addition, we implement an end-to-end REF toolbox for the deeper research in multi-modal domain. Our dataset and code can be available at: \url{https://qiuheqian.github.io/datasets/refcrowd/}. 
\end{abstract}

\ccsdesc[500]{Computing methodologies~Natural language processing}
\ccsdesc[500]{Computing methodologies~Computer vision tasks}
\ccsdesc[500]{Computing methodologies~Object detection}
\keywords{crowd understanding, referring expression comprehension, fine-grained multi-modal attribute contrastive}


\maketitle

\section{Introduction}
As increasing world population and rapid development of urbanization, the crowd has frequently appeared in the real world such as various markets, stations, stadiums and so on. Thus, crowd understanding has much practical significant and is becoming an important research direction. A lot of efforts have been made in vision domain, including crowd detection \cite{Zhang_2021_CVPR,wen2021detection,huang2020nms}, crowd counting \cite{liu2021visdrone,Xu_2021_ICCV,Liu_2021_ICCV}, crowd tracking \cite{sundararaman2021tracking,ning2020lighttrack}, etc. However, there is no research to explore crowd understanding in multi-modal domain, which jointly comprehends vision and linguistic information and naturally bridges the intelligent agents communicating with human about the physical world. In this paper, we are the first to focus on crowd understanding in a representative multi-modal task, i.e., referring expression comprehension.

\begin{figure}
	\centering
	\includegraphics[angle=0,scale=0.5]{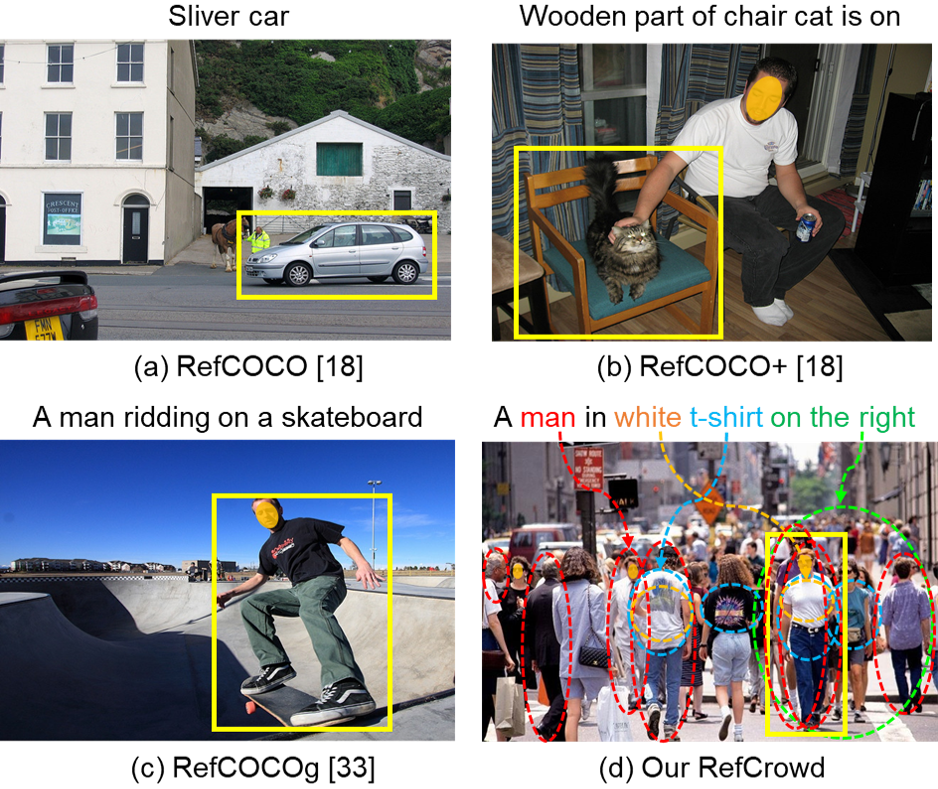}
	\vspace{-4mm}
	\caption{Some examples of comparison between previous REF datasets \cite{kazemzadeh2014referitgame,mao2016generation} and our RefCrowd dataset. Unlike previous datasets for general REF task, our Refcrowd focuses more on querying the target person in crowd, which requires to understand and distinguish only subtle differences between the target and similar persons according to the input expression. }
	\label{fig1} 
\end{figure}
Referring expression comprehension (REF) aims to locate a particular object in an image according to its referring expression with natural language, which plays an essential role in artificial intelligence research, including smart surveillance \cite{brutzer2011evaluation,shorfuzzaman2021towards}, security management \cite{moeslund2001survey,candamo2009understanding}, object of interest search \cite{zhang2018deep,li2017person} and human machine interactions \cite{norman1984stages,DBLP:journals/thms/GuoLY21}. However, a number of benchmark datasets \cite{kazemzadeh2014referitgame,mao2016generation,liu2019clevr,yang2020graph,chen2020cops,wang2020give} usually query and ground the target object from a variety of distinct object categories in general scenarios, such as the car, cat and person as shown in Fig. \ref{fig1}. Although they have involved person grounding, the crowd in real world is under-represented in these datasets. According to our statistics, the average number of persons per image is only 2 to 5 on typical REF datasets, e.g., RefCOCO \cite{kazemzadeh2014referitgame}, RefCOCO+ \cite{kazemzadeh2014referitgame} and RefCOCOg \cite{mao2016generation}. Since the crowd in real world contain a large amount of persons who look similar in appearance and occlude with each other, which makes more challenging to ground the target person with referring expressions. For instance, it is easy to correctly locate the target person only using the category information in Fig. \ref{fig1} (c) from RefCOCOg dataset \cite{kazemzadeh2014referitgame}, while it can lead to confusion due to the presence of multiple persons in Fig. \ref{fig1} (d) in crowd.

To move forward the field of REF, we propose a new challenging dataset, RefCrowd, 
to ground the person in crowd with referring expressions. Our dataset contains a crowd of persons some of whom share similar visual appearance, and diverse natural languages covering unique properties of the target person. It not only requires to sufficiently mine and understand natural language information, but also requires to carefully focus on subtle differences between persons in an image, so as to realize fine-grained mapping from language to vision. In our RefCrowd dataset, the statistic average number of persons per image reaches 16.8, which far exceeds existing datasets. Meanwhile, there are rich person attributes in language expressions and a variety of real crowd scenes. We further provide a detailed statistic analysis on our dataset and compare with existing datasets in terms of images, expressions and number of persons.

In order to deal with REF in crowd understanding, we further propose a one-stage Fine-grained Multi-modal Attribute Contrastive Network (FMAC) for fine-grained attribute features matching between the language and vision. Specifically, we use an attribute-aware multi-modal decomposition module (AMD) to decompose intricate features of language and vision into explicit attribute-level features, and then integrate language and visual features at the attribute level. 
According to parsed attributes from the language expressions, each type of attribute contains multiple fine-grained attribute classes, e.g., the gender attribute type includes male and female. Based on the attribute-aware multi-modal features, we design a fine-grained attribute contrastive module (FAC) to capture discriminative 
fine-grained attribute features and then leverage them to distinguish the described target from a crowd of persons with similar attributes. The fine-grained attribute features in FAC are learned by an attribute contrastive loss and an attribute classification loss. The attribute contrastive loss is designed to push away different fine-grained attribute features within each attribute type, and keep the consistent with the same fine-grained attribute class in the whole dataset. The attribute classification is used to filter and suppress unrelated attribute features with the target person. Extensive experiments are conducted on our RefCrowd dataset and general datasets to demonstrate the effectiveness of our method. 


The main contributions of this paper are summarized as:

$\bullet$ We propose a new challenging dataset RefCrowd that aims at grounding the person in crowd with referring expressions, which is the first attempt to explore crowd understanding in multi-modal task. Comprehensive analysis prove the standardization and superiority of our dataset in REF of crowd understanding.

$\bullet$ We propose a Fine-grained Multi-modal Attribute Contrastive Network towards REF in crowd understanding, which focuses on fine-grained attribute mapping from language to vision. The proposed method outperforms existing state-of-the-art methods on our RefCrowd dataset and previous datasets.

$\bullet$ Instead of conventional two-stage REF methods, we implement a code toolbox based on open-source MMDetection \cite{mmdetection} for end-to-end REF task, which flexibly supports the integration of natural language and various visual detectors.
\vspace{-3mm}
\section{Related Work}
\setlength{\parindent}{2em}\textbf{Datasets.} 
A number of datasets have been constructed for referring expression comprehension task. An early study is ReferItGame dataset \cite{kazemzadeh2014referitgame}, which collected large-scale referring
expressions for real-word objects by a two-player online game. Following ReferItGame dataset, RefCOCO and RefCOCO+ datasets \cite{kazemzadeh2014referitgame} collected the images from MSCOCO dataset \cite{lin2014microsoft}, where RefCOCO dataset has no restrictions on the type of language while RefCOCO + focuses more on purely object appearance description than location. RefCOCOg dataset \cite{mao2016generation} provided longer and more complex sentences constructed by a non-interactive setting. Flickr30k Entities \cite{plummer2015flickr30k} built correspondence between phrases in sentences and regions in images. Furthermore, a few datasets have been proposed to evaluate the reasoning ability of the model, such as CLEVER-Ref \cite{liu2019clevr}, Cops-Ref \cite{chen2020cops}, Ref-reasoning \cite{yang2020graph} and KB-Ref \cite{wang2020give} datasets, etc. In this paper, we propose a RefCrowd dataset to fill the gap of REF in complex crowd understanding.
%

\textbf{Approaches.} 
Recent approaches in referring expression comprehension can be basically divided into two categories, two-stage methods and one-stage methods. Two-stage methods \cite{luo2017comprehension,niu2019variational,wang2019neighbourhood,yu2018mattnet,hu2017modeling,liu2019improving,yang2019dynamic,yang2019cross,liu2020learning,liu2019learning,cirik2018using,cheng2021exploring} reformulated REF as a retrieval problem of region-language pairs.
In the first stage, a set of region proposals are generated relied on a pre-trained object detection network \cite{ren2015faster,he2017mask}. In the second stage, the best matching region is selected according to the rank of similarity between these proposals and a language query. 
These methods usually focus more on improving the second stage by joint embedding \cite{luo2017comprehension,niu2019variational,wang2019neighbourhood}, modular attention networks \cite{yu2018mattnet,hu2017modeling,liu2019improving}, object relational reasoning \cite{yang2019dynamic,yang2019cross,liu2020learning} and parsed language-guided learning \cite{liu2019learning,cirik2018using}. 

Since the overall performance of two-stage methods is inevitable capped by the quality of region proposals in the first stage, recent one-stage methods \cite{yang2019fast,yang2020improving,sun2021iterative,qiu2020language,liao2020real,deng2021transvg,li2021bottom,huang2021look} get rid of the limitations and enable end-to-end joint optimization. Instead, these one-stage methods directly perform the bounding box prediction based on fused multi-modal features of language and vision. A pioneering work FAOA \cite{yang2019fast} embedded the language features into the YOLOv3 \cite{redmon2018yolov3} object detector to ground the referred object, as do ReSC \cite{yang2020improving} and LBYLNet \cite{huang2021look}. RCCF\cite{liao2020real} treated the language domain as a kernel and performed correlation filtering on CenterNet \cite{duan2019centernet} to predict the object center. TransVG \cite{deng2021transvg} established a multi-modal transformer-based framework and grounded objects by directly regressing coordinates. 

The above methods usually understand language at the whole sentence or literal word level, and match it with object instances or fixed image regions. Although they have achieve advanced performance in general scenarios, it is insufficient for challenges of complex real-world scenarios. To address this problem, we bridge language and visual features at the fine-grained and representative person attribute level.

\textbf{Tasks.} 
There are two person-centric tasks related to the one we propose. One task is crowd detection \cite{shao2018crowdhuman,zhang2019widerperson,zhang2017citypersons} that aims to detect all persons in a crowd image. However, it only involves single vision modality. Although another text-based person ReID task \cite{wang2017adversarial,zhu2021dssl,li2017person} includes the language and vision domain, it ignores real-world complex background and only contains a cropped person for each image. This task expects to retrieve the cropped person images from the whole dataset given the input text, which focuses on the directly conversions between modalities (e.g., Text$\rightarrow$Image, Image$\rightarrow$Text) and determine whether these belong to the same identity, instead of scene understanding. Unlike the
above two tasks, our task aims to locate the described person according to the input crowd image and expression (Image+Text$\rightarrow$person), which requires to further understand the context of text and crowd scene.
\begin{table*}[htbp]
	\centering
	\caption{Comparison of different referring expression comprehension datasets in person grounding.}
	\vspace{-3mm}
	\scalebox{0.83}{
		\setlength{\tabcolsep}{3mm}{
			\begin{tabular}{@{}l|c|c|c|c|c|c|c@{}}	
				\toprule	
				Dataset        & \makecell[c]{Images} & \makecell[c]{Expressions}  & \makecell{Queried Persons} &\makecell[c]{Persons \\per image}&  \makecell[c]{Queried persons \\ per image} & \makecell[c]{ Expressions \\ per image}& \makecell[c]{ Avg. Expressions \\ Length }  \\
				\midrule
				RefCOCO \cite{kazemzadeh2014referitgame}        &9,969           &71,128             & 24,767                                   &5.23                  &  2.48                                           & 7.13                     &9.82                    \\
				RefCOCO+ \cite{kazemzadeh2014referitgame}      &9,969           &72,110               &24,752                                  &5.23                 &2.48                                      &\textbf{7.23}                &9.68                 \\
				RefCOCOg \cite{mao2016generation}      &10,186          &35,965              &18,304                                  &2.64                  &1.80                                     & 3.63                   &\textbf{17.81}                     \\
				RefCrowd (Our) & \textbf{10,702}           & \textbf{75,763}      &\textbf{37,999}                                   &\textbf{16.8}                  &\textbf{3.50}                                       &7.08        &13.13  \\
				\bottomrule	
	\end{tabular}}}
	\label{comparison dataset}
	\vspace{-2mm}
\end{table*}

\section{RefCrowd Dataset}

Different from previous REF datasets that mainly cover multiple distinctive object categories in general scenarios, we introduce a new dataset RefCrowd to ground the person in crowd with referring expressions. In this section, we first describe the pipeline of the dataset collection and annotation, and then provide a informative statistics analysis of our dataset.

\subsection{Dataset Collection}

\textbf{Images Collection.} 
We collected images of our dataset from popular MS COCO dataset \cite{lin2014microsoft} and open Internet. 
Based on annotations of the MS COCO dataset, we can conveniently sample 5,560 crowd scene images with multiple persons. In order to approach the diversity of real world scenarios, similar to \cite{plummer2015flickr30k,krishna2017visual,shao2018crowdhuman}, we further collected images from open image search engine (Google or Bing) with common scene keywords for query (e.g., sports, street, station, shopping, park, supermarket, classroom, etc.). To ensure the balance of image scene distribution, we restricted the number of images for each keyword to 300. Moreover, we filtered out some images with the small number of persons and unethical images. After those filter process, 5,142 images from Internet are remained. Finally, the whole dataset contains 10,702 images with crowd scenes.

\begin{figure}[htbp]
	\centering
	\includegraphics[angle=0,scale=0.33]{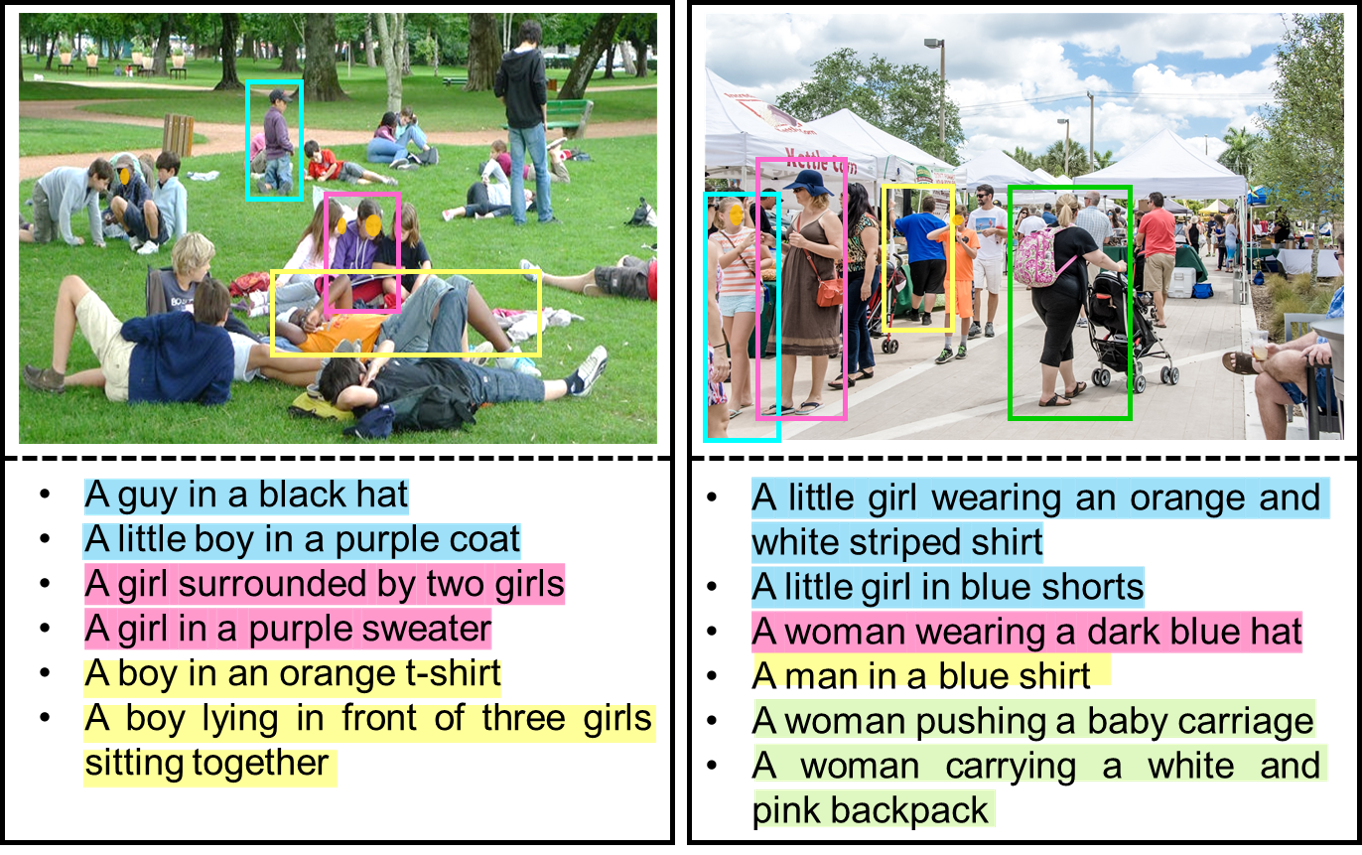}
	\vspace{-2mm}
	\caption{Illustrative examples from our RefCrowd dataset. In each image, there are multiple target persons (labeled by bounding boxes with different colors) described by expressions (highlighted by corresponding box colors). Each target person is labeled by diverse expressions.}
	\label{fig2} 
\end{figure}

\textbf{Annotations Collection.} 
The annotations of our dataset involve referring expression annotation and location annotation of the queried person in an image, as shown in Fig. \ref{fig2}. To ensure the quality of dataset, instead of Amazon Turk, we employed a professional team of 24 workers and spent nearly half a year to collect the dataset. During the annotation process, the following requests are put forward to ensure the effectiveness and quality of annotations. 1) A unified annotation python tool LabelRef is implemented for annotators to simultaneously label referring expression annotations and location annotations. 2) The location of queried person is tightly bounded by a rectangular bounding box. 3) In the same image, the written referring expressions of different persons are required to cover the unique attribute properties so as to correctly localize the queried person and avoid the ambiguity with other non-target persons. 4) If the same person is labeled with multiple expressions, they are required to contain different attributes for the diversity of annotations. 5) It is worth nothing that all expressions are forbidden to violate any ethical principles, including privacy, personal attack and impact social order. 

\textbf{Quality Control.} After labeling all annotations of the dataset following the above requirements, we abide by the principle of cross-checking to strictly check the quality of dataset. Specifically, we invited 24 examiners and divided them into 4 groups to performs cross-checking in groups. We firstly asked them to check whether all data come from public scenes to avoid privacy concerns and remove potentially unethical images and annotations. Then, we asked them to check erroneous annotations, such as without following the labeling requirements, misalignment of location and expression, semantic ambiguity, etc. In addition, we adopted a Python LanguageTool language-check as syntactic objective evaluations to conform to basic syntactic rules. Erroneous annotations would be corrected until they pass the check. In each group, each expression was checked at least 3 persons to avoid the subjective and noisy. Finally, we invited 5 professional inspectors to conduct a comprehensive supervision and review for further quality control. 

\textbf{Ethical Considerations.} Our dataset was constructed with careful consideration and examination to ethical issues including image collection and expression labeling. Researchers requires to sign RefCrowd Terms of Use as restrictions on access to dataset to privacy protection and use dataset for research purpose only. In addition, we allow people contact us to make reasonable suggestions. We masked all faces details in this paper.
\begin{figure}[htbp]
	\centering
	\includegraphics[angle=0,scale=0.25]{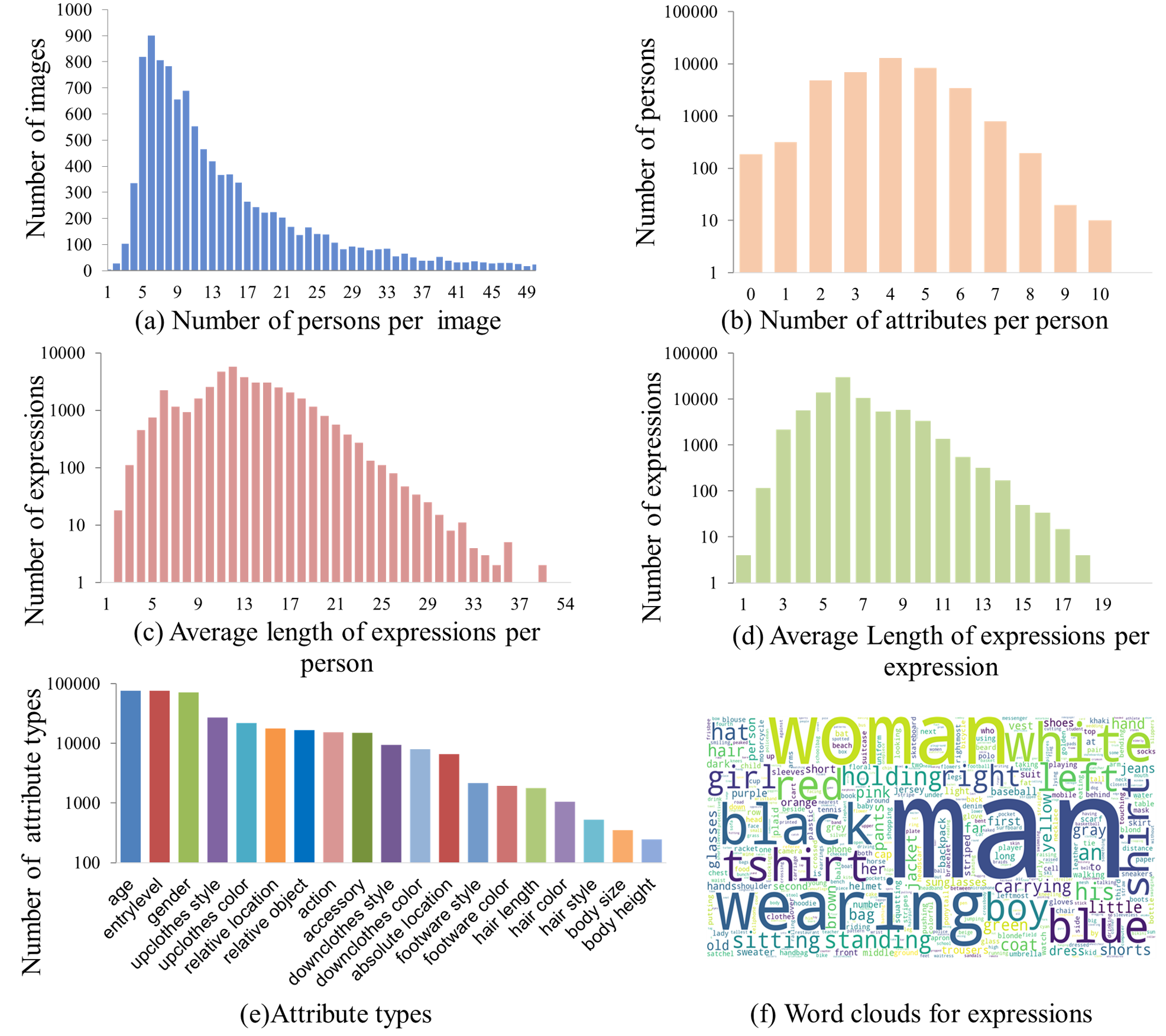}
	\vspace{-3mm}
	\caption{Statistical analysis of the proposed RefCrowd dataset, including the distribution of number of persons per image, attributes per person, average length of expressions per person and per expression, attribute types and word clouds for expressions in (a), (b), (c) and (d), (e) and (f).}
	\label{fig3} 
\end{figure}
\subsection{Dataset Stastics}\label{dataset}
Our dataset contains 75,763 expressions for 37,999 queried persons with bounding boxes on 10,702 images. Following random selection, we split the entire dataset into 6,885 images with 48,509 expressions for training, 1,260 images with 9,074 expressions for validation and 2,557 images with 18,180 expressions for testing, respectively.  

\begin{figure*}
	\centering
	\includegraphics[angle=0,scale=0.35]{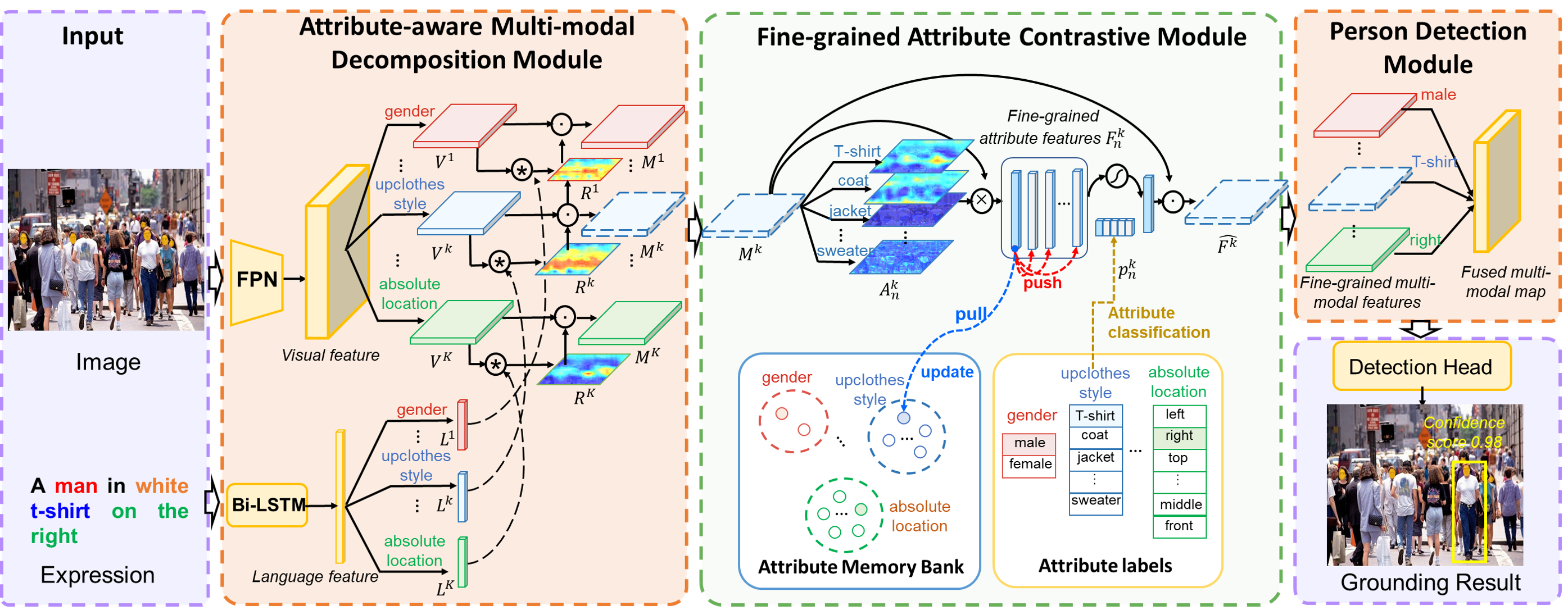}
	\vspace{-3mm}
	\caption{The overall architecture of our proposed Fine-grained Multi-modal Attribute Contrastive Network (FMAC), including attribute-aware multi-modal decomposition module (ADM), fine-grained attribute contrastive module (FAC) and person detection module. In our method, FAC is implemented for each attribute-aware multi-modal feature map. We highlight one type of attribute (i.e., uplothes style) and its corresponding fine-grained attributes in this figure, and other types of attribute are similar to it.}
	\label{network} 
\end{figure*}

Table \ref{comparison dataset} compares the statistics in person grounding of our RefCrowd dataset and other common referring expression datasets including RefCOCO \cite{kazemzadeh2014referitgame}, RefCOCO+ \cite{kazemzadeh2014referitgame}, RefCOCOg \cite{mao2016generation}. As shown in Table \ref{comparison dataset}, the scale of our dataset consistently surpasses other datasets in person grounding in terms of the total number of images, expressions and queried persons, especially compared with RefCOCOg \cite{mao2016generation}, the expressions and queried persons are more than twice. In order to verify the density of persons, we utilize a trained detector \cite{lin2017feature} on popular CrowdHuman dataset \cite{shao2018crowdhuman} to evaluate the number of persons. On average, there are 16.8 persons per image in our dataset, which is higher crowdness compared to other datasets with only a few persons. Due to the lower crowd density on other datasets, they are insufficient to serve as a ideal benchmark for the challenging crowd understanding. The corresponding crowdness distribution of our dataset is shown in Fig. \ref{fig3} (a). It can be observed that most of images include multiple person range from 5 to 25. To measure the standardization of expressions, we statistic the word length of our dataset on per person and per expression as shown in Fig. \ref{fig3} (c) and (d). The expression length is approximately lied in the range of 3 to 27 words for per person, and in the range of 2 to 17 words per expression. Averagely, there are 13.13 words for per person, which exceeds RefCOCO  \cite{kazemzadeh2014referitgame} and RefCOCO+ \cite{kazemzadeh2014referitgame}, where RefCOCOg \cite{mao2016generation} is built specifically for long expressions. Fig. \ref{fig3} (e) illustrates word clouds in our dataset, there are diverse words, such as man, woman, black, t-shirt, left, etc.

In order to make a further insight into our dataset, we use a Stanford CoreNLP parser \cite{socher2013parsing} to parse expressions and define 19 person attribute types similar to \cite{kazemzadeh2014referitgame}. Fig. \ref{fig3} (b) shows the distribution of number of attributes per person. Most of expressions per person involve the number of fine-grained attributes from 2 to 8, which demonstrates rich attributes often are required to describe the person in crowd. Fig. \ref{fig3} (e) shows their distribution of number of attribute types in our dataset. We can observe that our expressions cover diverse attributes, which is more applicable for the complex real-world scenarios and more challenging for this REF task. 

\section{Method}
To deal with crowd understanding in REF, we propose a Fine-grained Multi-modal Attribute Contrastive Network (FMAC) to locate the target person by fine-grained mapping from the language to vision. The overall architecture is illustrated in Fig. \ref{network}. Given an image and a language expression, we first decompose visual and language features into multiple attribute-level representations, and then map the attribute-aware language features to corresponding vision domain at each attribute level, respectively. Based on the attribute-aware multi-modal feature maps, we next perform fine-grained attribute contrastive to encourage the distinctiveness and robustness of attribute features for accurate person grounding. Finally, we employ a detection head to directly predict the target location and confidence score.

\subsection{Attribute-aware Multi-modal Decomposition Module}
In complex crowd understanding, it is no longer enough to distinguish persons with similar appearance by only using instance-level object information in previous methods \cite{yu2018mattnet,yang2020graph,huang2021look}. Based on the statistic in Section \ref{dataset}, we decompose intricate visual and language features into multiple specific person attributes to simplify the network learning of complex task, which includes visual attribute decomposition, language attribute decomposition and multi-modal attribute fusion between them. 

In visual attribute decomposition, we adopt ResNet \cite{he2016deep} based on Feature Pyramid Network (FPN) \cite{lin2017feature} as image encoder to extract visual features from different scales. The visual feature map of a certain scale is represented as $V\in \mathbb{R} ^{C \times H \times W}$. Then, we split the visual feature map into $K$ types using different $1\times1$ convolution layers and obtain $K$ feature maps $\{V^k\}_{k=1}^{K}$, where $V^k\in \mathbb{R} ^{C_k \times H \times W}$ denotes the visual feature map corresponding to the $k$-th attribute type. In language attribute decomposition, we first embed each word of the input language expression into a one-hot vector, and then employ a Bi-LSTM to encode them sequentially. The last hidden states of its forward and backward are concatenate as the language feature $L\in\mathbb{R}^{C_q}$. Similar to visual features, we also split the language feature $L$ into $K$ types of attribute features $\{L^k\}_{k=1}^{K}$,  $L^k \in \mathbb{R} ^{C_k}$, and treat language attribute features as a set of attribute-aware filters $L^k$. Then, we convolve visual attribute features with corresponding language filters to obtain their semantic relation map $R^k \in \mathbb{R} ^{1 \times H \times W}$. Based on the semantic relation map, we can fuse the weighted feature map and original visual feature map at each attribute type to generate the final attribute-aware multi-modal feature map $M^k \in \mathbb{R} ^{C \times H \times W}$, in which original visual feature map $V^k$ is used to compensate for the information that may be lost after weighting. The specific calculation is as follows: 
\begin{equation}\label{multi-modal}	
\begin{split}
L^k&=tanh(W_qL), \\
R^k&=Softmax[L^k\ast V^k] ,\\
M^k&=W_m[R^k \odot V^k;V^k]
\end{split}
\end{equation}
where $W_m$ and $W_q$ are learnable weight parameters. $\ast$, $\odot$ and $[;]$ represent convolution, hadmard product and concatenate operators, respectively. Different attribute types usually focus on different spatial regions or semantic information. For example, the attribute map of gender focus on the global semantic features while the upclothes style focuses on the upper local part of the body.

\subsection{Fine-grained Attribute Contrastive Module}
It is difficult to distinguish different fine-grained attributes within the same attribute type due to their similar appearance and semantic information, e.g., there is subtle length difference between T-shirt and coat in the type of upcothes style (the style of upper body clothes). Based on multi-modal attribute feature map $M^k$ of each type, we first generate $N$ attention maps $\{A_n^k\}_{n=1}^{N}$ using a convolution layer, where each attention map $A_n^k \in \mathbb{R}^{1\times H\times W}$ represents related response regions corresponding to each fine-grained attribute.  $N$ is the number of fine-grained attributes at the $k$-th attribute type. Then, we can gather their representative semantic feature vector $F_n^k \in \mathbb{R}^{C}$ from each attribute map $M^k$ according to the response regions.
\begin{equation}\label{key}	
\begin{split}
F_n^k&=A_n^k \times {M^k}^T,\\
A_n^k&=Softmax[W^k_aM^k] ,
\end{split}
\end{equation}
where $W^k_a$ is weight parameter. $T$ denotes the transpose operation, $F_n^k$ is the feature vector of the $n$-th fine-grained attribute in the $k$-th attribute type.

To enhance the distinctiveness between different fine-grained attributes within the same attribute type, we design an attribute contrastive loss function $L_{ACO}$ and an attribution classification loss function $L_{ACE}$ to optimize the learning of these features. The goal of attribute contrastive loss function is to push away features of different fine-grained attributes while pull closer to the same one in the entire dataset. The attribution classification loss is used to constrain the attribute category to which the target person belongs. 

Suppose $l^k$ denotes the category label of ground-truth attribute at the $k$-th attribute type. $F_l^k\in \mathbb{R}^{C}$ denotes the semantic feature vector corresponding to the ground-truth label. $F_{n\neq l}^k\in \mathbb{R}^{C}$ represents the semantic vector of other negative categories in the same image and expression. Since there are different views for the same fine-grained attribute in different images, we maintain their consistency to make the robustness of features. Inspired by \cite{he2020momentum}, we build a memory bank $Q^k\in \mathbb{R} ^{N\times C}$ to store their semantic features with the same fine-grained attribute in entire dataset, where $Q_l^k$ denotes memory features for the $l$-th fine-grained attribute at the $k$-th type. The attribute contrastive loss $L_{ACO}$ can be defined as:
\begin{equation}
{L_{ACO}}_n^k=-log \frac{exp(F^k_l \cdot Q^k_n / \tau)}{exp(F^k_l \cdot  Q^k_n /\tau)+\sum_{n\neq l} exp(F^k_l \cdot F^k_n /\tau)},
\end{equation}
where $\tau=0.2$ is a temperature constant similar to \cite{he2020momentum}. Unlike previous triplet loss \cite{yu2018mattnet} that is limited to the number of positive and negtive samples, we consider all negative fine-grained attribute samples in current expression and positive sample with the same attribute across expressions and images. This helps to enhance the discrimination without losing the robustness of each attribute feature. During training, we first random initial the memory bank $Q^k$. Then, we update the value in $Q^k$ using moving average after each iteration:
\begin{equation}
Q_l^k \leftarrow mQ_l^k +(1-m)F_l^k,
\end{equation}
where $m=0.999$ represents the momentum.   

In addition, we predict their probability score $p_n^k$ for the described person to measure the weights of these attribute features, which can be directly supervised by an attribute classification cross entropy loss $L_{ACE}$:
\begin{equation}
\begin{split}
L_{ACE}&=-\sum_{n=1}^{N}w_n^k[y_n^klog(p_n^k)],\\
p_n^k&=\frac{exp(f_n^k)}{\sum_{n=1}^{N}exp(f_n^k)}, \\
\end{split}
\end{equation}
where $w_n^k=1/\sqrt{{Freq_{attr}}_n^k}$ weights the attribute labels to alleviate the imbalance problem of data, ${Freq_{attr}}_n^k$ is the frequency of persons with the attribute label $n$. $y_{n=l}^k=1$ when $n=l$ is the ground-truth label. Otherwise, $y_{n\neq l}^k=0$. Here, we use a fully connect layer to generate the probability logits $f_n^k\in \mathbb{R}^N$ and a softmax function to obtain the final probability score $p_n^k$, where a high probability score means the target person is more likely to contain this attribute category.

According to the predicted probability score, we can choose its representative fine-grained semantic feature in each attribute type by softly weighting these attribute features. Thereby, the unrelated attribute category features are suppressed while the important one is reserved. Then, we leverage the fused fine-grained attribute feature $\widehat{F^k}\in \mathbb{R}^C$ to enhance the initial decomposed attribute features $M^k$.
\begin{equation}\label{key}	
\begin{split}
\widehat{F^k}&=\sum_{n=1}^{N}p_n^kF_n^k,\\
\widehat{M^k}&=\sigma(\widehat{F^k})\odot M^k +M^k,\\
\end{split}
\end{equation}
where $\sigma$ denotes Sigmoid activate function, which scales the feature values of $\widehat{F^k}$ to the range of $[0,1]$.  Further, we concatenate these decomposed features and employ a convolution layer to generate the final fine-grained multi-modal feature map $\widehat{M}$.

\subsection{Person Detection Module}
Based on the captured fine-grained multi-modal feature map $\widehat{M}$, we adopt an anchor-free detection head FCOS \cite{tian2019fcos} including classification and localization branches to locate the target person corresponding to the input expression. For each location, we separately use four convolution layers to predict a 4D vector bounding box coordinates $t=(l,t,r,b)$ for the distance from the location to the four side of the ground-truth bounding box $t^*=(l^*,t^*,r^*,b^*)$, and a 1D vector the confidence score $c$ whether it belongs to the target person, $c^*$ is the ground-truth label. The detection head is optimized by a regression loss $L_{reg}$ and a classification loss $L_{cls}$ as follows:
\begin{equation}\label{key}	
L_{Det}=L_{reg}(t,t^*)+L_{cls}(c,c^*),
\end{equation}
where the regression loss $L_{reg}$ is GIoU loss \cite{rezatofighi2019generalized} and the classification loss $L_{cls}$ is focal loss \cite{lin2017focal}. 

To sum up, the overall network is end-to-end optimized by the aforementioned loss functions (3), (5) and (7) for the attribute features learning and the detection loss function:
\begin{equation}\label{key}	
L_{all}=L_{Det}+L_{ACO}+L_{ACE}.
\end{equation}

\section{Experiments}


\textbf{Implement Details.} We implement an end-to-end referring expression grounding toolbox to flexibly support the 
integration of natural language and various popular detectors based on open source MMDetection \cite{mmdetection} toolbox. Unless specified, all experiments of our method adopt a recent representative anchor-free detector FCOS \cite{tian2019fcos} with ResNet101-FPN \cite{lin2017feature}. The overall architecture of the proposed FMAC method is end-to-end optimized using SGD optimizer with a batch size of 16 for 12 epochs. We set the base learning rate to 0.02 and decrease by a factor of 10 after 8 epochs and 11 epochs. Because the visual encoder and detection head are initialized using pre-trained on the MS COCO dataset \cite{lin2014microsoft}, we multiply their learning rate by 0.1. The scale of input image is resized to $800\times1333$, following the default settings of FCOS \cite{tian2019fcos}. It is worth mentioning that we do not use any data augmentation throughout the training and inference stages. The channels of encoded language, visual and attribute feature are set to $C_q=1024$, $C=256$ and $C_k=256$. The number of attribute types is set to $N=8$.

\textbf{Evaluation Metric.} We calculate the intersection-over-union (IoU) between the predicted bounding box and ground-truth one to measure whether the prediction is correct. A predicted bounding box is treated as correct if IoU is higher than desired IoU threshold. Instead of using single IoU threshold 0.5, we adopt the mean accuracy $mAcc$  to measure the localization performance of methods, which averages the accuracy over IoU thresholds from loose 0.5 to strict 0.95 with interval 0.05 similar to popular COCO metrics \cite{lin2014microsoft}. $mAcc$ is a comprehensive indicator for widely real-world applications.


\subsection{Dataset Bias Analysis}
Analyzing the impact of dataset bias is necessary to the further research on vision and language. Inspired by \cite{cirik2018visual}, we shows the results of dataset bias analysis in Table \ref{bias}. The \emph{Random} predicts a random bounding box based a pre-trained object detector FCOS, which only achieves 6.90\% on validation set and 5.10\% on test set in term of loose $Acc_{50}$. Without the language expression, the \emph{w/o Expressions} only uses the image to train the model and obtains better performance than \emph{Random}. Introducing partial expressions with the subject or all nouns, we observe the performance of \emph{Subject Expressions} has barely improved compared to the \emph{w/o Expressions}, while there is still a large gap between the performance of \emph{Partial Expressions} and our method. 
This result demonstrates our RefCrowd dataset is more
challenging and only using partial expressions is not enough, 
which requires the algorithm carefully understands more expression information so as to correctly locate the target because a crowd scene usually contains multiple persons related to the subject expression. 
\begin{table}[htbp]
	\centering
	\caption{Dataset bias analysis with different settings on the RefCrowd dataset. w/o means the input is removed.}
	\vspace{-2mm}
	\scalebox{0.83}{
		\setlength{\tabcolsep}{2mm}{
			\begin{tabular}{l|lll|lll}
				\toprule
				\multicolumn{1}{l|}{\multirow{2}{*}{Method}} & \multicolumn{3}{c|}{val}                                    & \multicolumn{3}{c}{test} \\
				\cline{2-7} 
				\multicolumn{1}{c|}{}                        & \multicolumn{1}{l|}{$Acc_{50}$} & \multicolumn{1}{l|}{$Acc_{75}$} & $mAcc$ & \multicolumn{1}{l|}{$Acc_{50}$} & \multicolumn{1}{l|}{$Acc_{75}$} & $mAcc$ \\
				\midrule
				Random                                       & \multicolumn{1}{l|}{6.90}   & \multicolumn{1}{l|}{6.41}       &   5.79  & \multicolumn{1}{l|}{5.10}     & \multicolumn{1}{l|}{4.79}    &4.27      \\
				w/o Expressions                            & \multicolumn{1}{l|}{24.47}   & \multicolumn{1}{l|}{22.28}       & 20.27    & \multicolumn{1}{l|}{25.21}     & \multicolumn{1}{l|}{23.46}    & 21.03     \\
				Subject Expressions                       & \multicolumn{1}{l|}{24.92}   & \multicolumn{1}{l|}{22.75}       & 20.47    & \multicolumn{1}{l|}{25.26}     & \multicolumn{1}{l|}{23.37}    & 20.98 \\ 
				Partial Expressions                       & \multicolumn{1}{l|}{45.36}   & \multicolumn{1}{l|}{40.22}       & 36.07    & \multicolumn{1}{l|}{45.12}     & \multicolumn{1}{l|}{40.19}    & 35.88 \\ 	
				Our FMAC                                         & \multicolumn{1}{l|}{\textbf{57.32}}   & \multicolumn{1}{l|}{\textbf{50.66}}       &   \textbf{45.51}  & \multicolumn{1}{l|}{\textbf{57.47} }     & \multicolumn{1}{l|}{\textbf{50.81}}    & \textbf{45.58}     \\ 
				\bottomrule
	\end{tabular}}}
	\label{bias}
	\vspace{-3mm}
\end{table}
\begin{table}[htbp]
	\centering
	\caption{The effects of main components in the proposed method on the RefCrowd validation set.}
	\vspace{-2mm}
	\scalebox{0.83}{
		\setlength{\tabcolsep}{2.3mm}{
			\begin{tabular}{@{}c|c|c|c|c|c|c@{}}	
				\toprule	
				ADM&FAC&FAC+$L_{ACO}$&FAC+$L_{ACE}$&$Acc_{50}$&$Acc_{75}$&$mAcc$\\
				\midrule	
				&	&&&53.77&47.42&42.71\\
				\checkmark&&&&56.40&49.74&44.85\\ 
				\checkmark&\checkmark&&&56.08&49.90&44.68\\ 
				\checkmark&\checkmark&&\checkmark&56.39&49.99&44.90\\
				\checkmark&\checkmark&\checkmark&&57.05&50.28&45.25\\
				\checkmark&\checkmark&\checkmark&\checkmark&\textbf{57.32}&\textbf{50.66}&\textbf{45.51} \\
				\bottomrule	
	\end{tabular}}}
	\label{main componets}
	\vspace{-2mm}
\end{table} 

\subsection{Ablation Studies}
\sloppy{}
\textbf{Effects of main components.} The effects of main components in our method are shown in Table \ref{main componets}. The baseline model directly fuses the language and visual features using Eq.\ref{multi-modal}. Compared with the baseline method, the attribute-aware multi-modal decomposition module (AMD) consistently improves the performance by 2.63\% and 2.14\% in terms of widely-used $Acc_{50}$ and comprehensive $mAcc$, which demonstrates the effectiveness of decomposed features fusion. In addition, we introduce the fine-grained attribute contrastive Module (FAC) without constraint loss function, the performance is slightly decreased. Significantly, the $mAcc$ of performance is improved when introducing the attribute contrastive loss $L_{ACO}$ and the attribute classification loss $L_{ACE}$,
especially for $L_{ACO}$ that improves the performance by nearly 1\%. It reveals that discriminative features are important to distinguish the differences between different attributes with similar semantic. The overall method significantly outperforms baseline by 3.55\%, 3.24\% and 2.8\% based on $Acc_{50}$, $Acc_{75}$ and $mAcc$, respectively. This results mean that it is necessary to capturing fine-grained attributes for REF in crowd understanding.

\textbf{Effects of different attributes.} We analyze the effects of different attributes in Table \ref{attribute types}. Following the most frequent strategy, we set the number of attribute type to $K=1,3,5,8,10$ respectively, and select top 86 fine-grained attribute categories in total. It can be observed that the performance is improved gradually with the increase of number K of attribute types and achieves the best performance when $K=8$ (i.e., these attribute types are entrylevel, gender, upclothes style, upclothes color, relative location, relative object, action and accessory, according to our statistics in Fig. \ref{fig3} (e). Because the age attribute type is often contained in entrylevel, we remove the type.). The performance is relatively declined when K=10. One possible reason is the potential data imbalance caused by the mixing of low-frequency attribute types. In addition, we fix the number of attribute type $K=8$. It can be observed that the frequent strategy for choosing attributes outperforms the random strategy by 0.67\% $mAcc$. Based on the above analysis, we set to K=8 using the frequent strategy. 

\begin{table}[htbp]
	\centering
	\caption{The effects of different attributes on the RefCrowd validation set.}
	\vspace{-3mm}
	\scalebox{0.83}{
		\setlength{\tabcolsep}{5mm}{
			\begin{tabular}{@{}l|c|c|c@{}}	
				\toprule	
				Method&$Acc_{50}$&$Acc_{75}$&$mAcc$\\
				\midrule	
				K=1&55.75&48.98&44.09\\
				K=3&56.19&49.28&44.23\\
				K=5&56.84&49.91&45\\
				K=8&\textbf{57.32}&\textbf{50.66}&\textbf{45.51}\\
				K=10&56.94&50.09&45.11\\
				\midrule
				Random attributes (K=8)&56.82&49.89&44.84\\
				Frequency attributes (K=8)&\textbf{57.32}&\textbf{50.66}&\textbf{45.51}\\
				\bottomrule	
	\end{tabular}}}
	\label{attribute types}
	\vspace{-4mm}
\end{table}
\begin{table}[htbp]
	\centering
	\caption{The effects of different contrastive loss function in FAC on the RefCrowd validation set.}
	\vspace{-2mm}
	\scalebox{0.83}{
		\setlength{\tabcolsep}{4mm}{
			\begin{tabular}{@{}l|c|c|c@{}}	
				\toprule	
				Method&$Acc_{50}$&$Acc_{75}$&$mAcc$\\
				\midrule	
				w/o attribute contrastive&56.39&49.99&44.90\\
				\midrule
				Triplet loss \cite{yu2018mattnet} &56.42&49.77&44.82\\
				Common contrastive loss \cite{he2020momentum} &56.59&50.03&44.85\\
				Our $L_{ACO}$ w/o Memory Bank &56.47&49.79&44.81\\
				Our $L_{ACO}$&\textbf{57.32}&\textbf{50.66}&\textbf{45.51}\\
				\bottomrule	
	\end{tabular}}}
	\label{contrastive loss}
	\vspace{-3mm}
\end{table}
\textbf{Effects of fine-grained attribute contrastive learning.} Table \ref{contrastive loss} studies the effects of fine-grained attribute contrastive learning. Triplet loss function randomly samples one positive and negative at a time, which is widely used in feature learning \cite{yu2018mattnet}. It can be observed that only one negative sample in triplet loss has loosed its validity in crowd with a large number of negative samples. Instead, our contrastive learning considers more attribute negative samples in current image and positive sample from memory bank with the entire dataset. This is also different from conventional contrastive learning \cite{he2020momentum} that treats the negative samples from the entire dataset, while it is more important to distinguish the objects in current image for REF task. In Table \ref{contrastive loss}, our attribute contrastive performs better than the two methods, which demonstrates the our attribute contrastive loss has more conductive to the distinguish features between persons in crowd.


\begin{table*}[htbp]
	\centering
	\caption{Comparison with state-of-the-art methods on our RefCrowd, RefCOCO \cite{kazemzadeh2014referitgame} and RefCOCO+ datasets \cite{kazemzadeh2014referitgame}.}
	\vspace{-2mm}
	\scalebox{0.8}{
		\setlength{\tabcolsep}{2.5mm}{
			\begin{tabular}{l|c|c|c|ccc|ccc|c|c}
				\toprule
				\multicolumn{1}{l|}{\multirow{2}{*}{Method}} & \multicolumn{1}{c|}{\multirow{2}{*}{Year}} & \multicolumn{1}{c|}{\multirow{2}{*}{Detector}} & \multicolumn{1}{c|}{\multirow{2}{*}{Backbone}} & \multicolumn{3}{c|}{RefCrowd val}                                                    & \multicolumn{3}{c|}{RefCrowd test}&\multicolumn{1}{c|}{RefCOCO}&\multicolumn{1}{c}{RefCOCO+ }                                                 \\ \cline{5-12} 
				\multicolumn{1}{c|}{}                        & \multicolumn{1}{c|}{}                      & \multicolumn{1}{c|}{}                          & \multicolumn{1}{c|}{}                          & \multicolumn{1}{c}{$Acc_{50}$} & \multicolumn{1}{c}{$Acc_{75}$} & \multicolumn{1}{c|}{$mAcc$} & \multicolumn{1}{c}{$Acc_{50}$} & \multicolumn{1}{c}{$Acc_{75}$} & \multicolumn{1}{c|}{$mAcc$}&\multicolumn{1}{c|}{testA} &\multicolumn{1}{c}{testA}    \\ 
				\midrule
				MattNet \cite{yu2018mattnet}                                      & 2018                                       & Mask R-CNN  \cite{he2017mask}                                   & ResNet-101                                     &       52.47                  &      45.60                   &       40.32                  &       52.50                  &      46.30                   &        40.66 & 81.14&71.62               \\
				CM-Att \cite{liu2019improving}                                      & 2019                                       & Mask R-CNN  \cite{he2017mask}                                   & ResNet-101                                     &       53. 22                 &      46.61                   &       41.00                  &       53.87                  &      47.51                   &        41.73  & 82.16&72.58                 \\
				CM-Att-Erase \cite{liu2019improving}                                      & 2019                                       & Mask R-CNN  \cite{he2017mask}                                   & ResNet-101                                     &       54.94                  &      48.01                   &       42.32                  &       54.98                  &      48.45                   &        42.50 & 83.14&73.65                 \\
				FAOA\cite{yang2019fast}                                          & 2019                                       & YOLOv3  \cite{redmon2018yolov3}                                      & DarkNet-53                                     &    42.34                     &     29.40                  &      26.60                   &      42.39                     &   29.62                      &         26.72 &74.35&60.23               \\
				ReSC-Large  \cite{yang2020improving}                                  & 2020                                       & YOLOv3   \cite{redmon2018yolov3}                                      & DarkNet-53                                     &      49.10                  &    39.72                     &    34.64                     &   49.55                      &     41.06                    &        35.16      &  80.45&68.36         \\
				LBYLNet  \cite{huang2021look}                                     & 2021                                       & YOLOv3     \cite{redmon2018yolov3}                                         & DarkNet-53                                     &     51.73                    &      37.47                   &     33.67                    &     52.40                    &          38.30              &        34.08  &82.18&73.38               \\
				TransVG   \cite{deng2021transvg}                                    & 2021                                       & DETR    \cite{carion2020end}                             & ResNet-101                                     &       42.51                  &   29.85                      &     27.57                    &        43.03                 &   30.43                      &     27.85   &82.72&70.70                 \\
				\bottomrule
				Baseline                                          & Our                                        & RetinaNet  \cite{lin2017focal}                                    & ResNet-101                                     &   43.67                      &       38.53                  &          34.22               &      45.21                 &  39.98                       &        35.52  &80.49             &70.43              \\
				Baseline                                          & Our                                        & Reppoint     \cite{yang2019reppoints}                                  & ResNet-101                                     &     53.9                     &       46.24                 &       41.11                  &    54.22                     &     47.11                    &      41.62               &  81.79   & 72.57  \\
				Baseline                                          & Our                                        & FCOS      \cite{tian2019fcos}                                     & ResNet-101                                     &      53.77                   &   47.42                      &    42.71                   & 54.79                        &  48.19                       &  43.15     &81.13&72.25                   \\
				
				\midrule
				FMAC                                          & Our                                        & RetinaNet     \cite{lin2017focal}                                  & ResNet-101                                     &        49.38                 &          43.39               &         38.75                &    49.82                     &   44.09                      &    39.28      &83.58  &72.77                \\
				FMAC                                          & Our                                        & Reppoint       \cite{yang2019reppoints}                                  & ResNet-101                                     &    \textbf{57.46}                      &    50.01                     &     44.37                    &   \textbf{57.80}                      &      50.25                   &    44.53      &\textbf{84.05}   &\textbf{75.02}               \\
				FMAC                                          & Our                                        & FCOS        \cite{tian2019fcos}                                   & ResNet-101                                     &      57.32                   &  \textbf{50.66}                      &    \textbf{45.51}                     &    57.47                    &     \textbf{50.81}                    &     \textbf{45.58} &83.50 &74.12                   \\
				\bottomrule
			\end{tabular}
	}}
	\label{comparison}
	\vspace{-2mm}
\end{table*}

\begin{figure}[htbp]
	\centering
	\includegraphics[angle=0,scale=0.18]{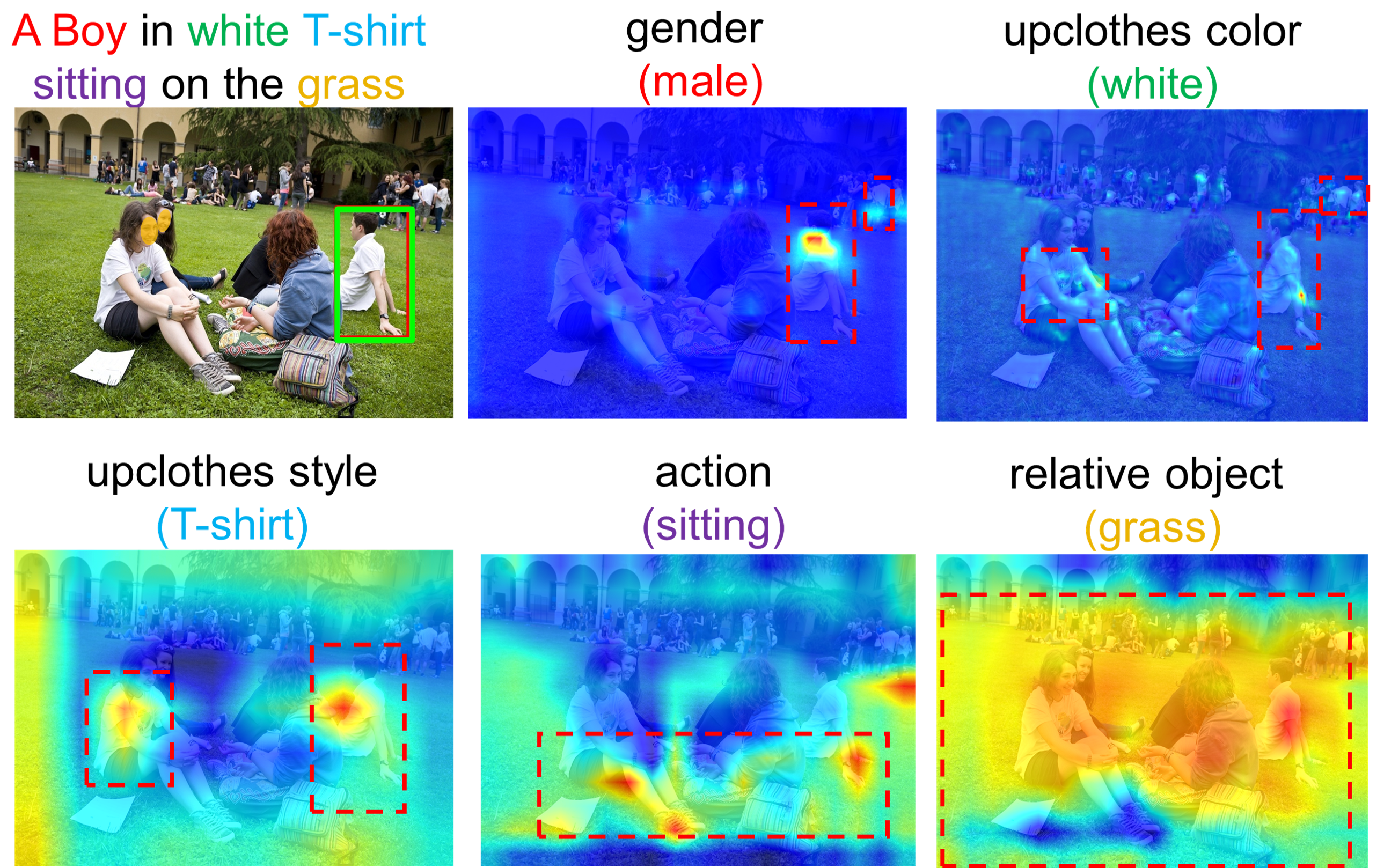}
	\vspace{-2mm}
	\caption{Visualization of attribute attention maps. The red dotted boxes show the highlighted regions.  }
	\label{attmap} 
	\vspace{-5mm}
\end{figure}

\textbf{Visualization Results.} Fig. \ref{attmap} show the attention maps of relevant attributes of grounding the target person in the inference process. It can be observed that relevant attribute areas of the query person are highlighted to assist fine-grained mapping from language to vision. For example, there are multiple persons sitting on the grass as shown in the attention map of ``grass". Some men are noticed for the``male" of gender attribute. The ``white" of upclothes color attribute and ``T-shirt" of upclothes style attribute highlight the local area of the upper body of persons. For the ``sitting" of action attribute, the legs and foot areas of these girls and hands of the boy on the ground is concerned. Combining these attention maps, it is naturally to find the particular person described by the expression from a crowd of persons with similar properties. 

\begin{figure}[htbp]
	\centering
	\includegraphics[angle=0,scale=0.31]{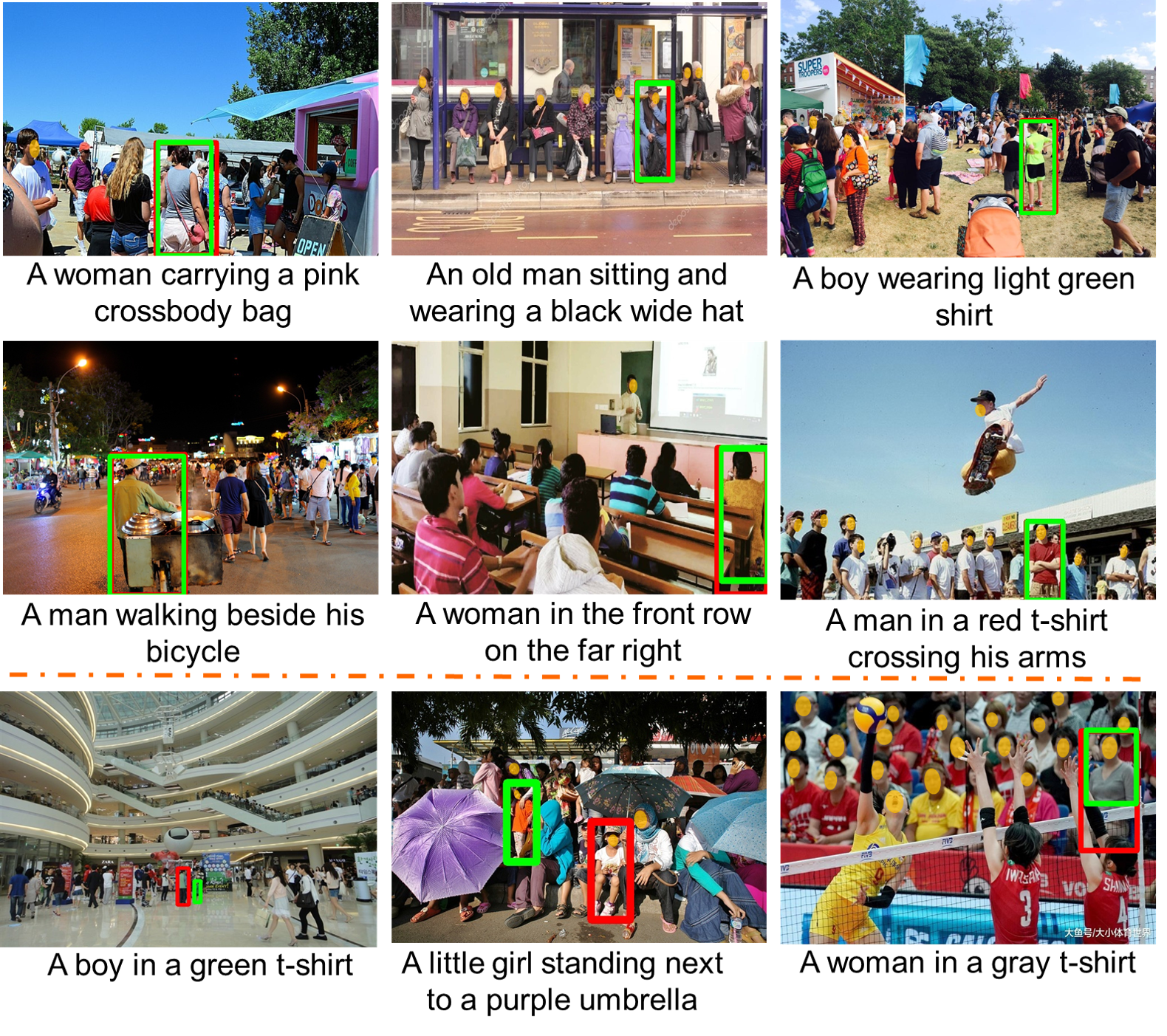}
	\vspace{-5mm}
	\caption{Visualization of grounding results on our RefCrowd dataset. The first two lines show some successful cases and the last line shows failure cases. The red and green boxes are the ground truth and the predict box.}
	\label{visualize} 
	\vspace{-6mm}
\end{figure}

More successful grounding cases are shown in the first two lines of Fig. \ref{visualize}. It demonstrates that the proposed fine-grained multi-modal attribute contrastive method is effective to accurate person grounding in more complex crowd understanding. Some failure cases are also reported in the last line of Fig. \ref{visualize}. It is difficult to locate the tiny target person, such as the boy and the man wearing the same clothes in the first case. Because the target persons are not visible, the disturbing objects similar to them would be incorrectly located with higher confidence score. In addition, complex and long expressions are always a challenge problem in REF task. For example, it is hard to distinguish the two girls both beside the umbrella in the second case. The third case shows a loose location result due to occlusion between person. In the future, we will attempt to introduce more informative features and language reasoning to address the tiny or occlusion object and complex language problems.

\subsection{Comparison with State-of-the-art Methods}
We compare the proposed method with state-of-the-art methods using various object detectors on our RefCrowd and general REF datasets in Table \ref{comparison}. For fair comparison, we train all methods on our RefCrowd dataset using their default strategies. Although existing methods have achieved advanced performance on general REF datasets, there is a large drop of their performance on our dataset, especially for one-stage methods. These results demonstrate the challenging of our dataset and inspire us to further propose our method FMAC to explore one-stage methods in complex scenarios. It can be observed that Our method based on FCOS \cite{tian2019fcos} suppresses all SoTA methods on validation and test sets of our RefCrowd dataset. In addition, we also implement our method based on other detectors using our REF toolbox, including anchor-based RetinaNet \cite{lin2017focal} and anchor-free Reppoint \cite{yang2019reppoints} detectors, which significantly outperforms the Baseline method that directly fuse the ambiguous visual and language features using Eq.\ref{multi-modal}. Meanwhile, our method outperforms these methods on testA set including person of general RefCOCO and RefCOCO+.The results demonstrate the generality and effectiveness of our method.

\vspace{-2mm}
\section{Conclusion}
In this paper, we have proposed a new challenging REF dataset, RefCrowd, which aims at looking for the persons in crowd based on referring expressions. It encourages the algorithm to leverage the natural language with unique properties of the target person, and carefully explore subtle differences from multiple persons with similar visual appearance. Based on this dataset, we have designed a fine-grained multi-modal attribute contrastive network to deal with REF in crowd understanding, which decomposes the coarse features into person attribute features, and captures fine-grained attribute matching between language and vision. Extensive dataset analysis and experiment results shows that the challenging and validity of our dataset and the effectiveness of the proposed method. To end up, we wish our dataset can attract more attention on crowd understanding for multi-modal domain in the future. 

\bibliographystyle{ACM-Reference-Format}
\bibliography{refcrowd}

\appendix

\end{document}